\documentclass[paragraphspacing,palatinofont]{nsfprop}

\newcommand{\ignore}[1]{}
\usepackage{graphicx}
\usepackage{wrapfig}
\usepackage{amsmath,amssymb}
\usepackage{amsmath}
\usepackage{algorithmicx}
\usepackage{algpseudocode}
\usepackage{cases}
\usepackage{wrapfig}
\usepackage{setspace}
\usepackage{tabularx}
\usepackage{enumitem}
\usepackage{url}
\usepackage{csquotes}
\usepackage{wrapfig}
\usepackage{pgfgantt}
\usepackage{color,hyperref}
\usepackage{caption}
\usepackage{subcaption}
 \usepackage{color,hyperref}
 \usepackage{times}
 \usepackage{xcolor}
 \usepackage{soul}
 \usepackage[utf8]{inputenc}
 \usepackage{amsthm}

 \theoremstyle{definition}
 
 \usepackage{todonotes}
 \usepackage{comment}

 \newcommand{\ms}[1]{{\textcolor{red}{~(MS: #1)}}}

\title{Declarative Learning-Based Programming as \\An Interface to AI Systems}
\begin{document}




\author{
 Parisa Kordjamshidi$^{a,b}$, 
 Dan Roth$^c$,
Kristian Kersting$^d$
 \\ 
 $^a$ Michigan State University, Dept. of Computer Science \& Engineering, USA\footnote{Parisa is currently affiliated with Michigan State University, however, this manuscript is prepared when working in Tulane University and IHMC.}\\
 $^b$ Florida Institute for Human and Machine Cognition, USA \\ 
 $^c$  University of Pennsylvania, Dept. of Computer and Information Science, USA\\
 $^d$  TU Darmstadt, Computer Science Dept.~\& Centre for Cognitive Science, Germany\\
{kordjams@msu.edu},
 {danroth@seas.upenn.edu},
 {kersting@cs.tu-darmstadt.de}
 }

\date{}
\maketitle

\begin{abstract}
Data-driven approaches are becoming more common as problem-solving techniques in many areas of research and industry.
In most cases, machine learning models are the key component of these solutions, but a solution involves multiple such models, along with significant levels of reasoning with the models' output and input. Current technologies do not make such techniques easy to use for application experts who are not fluent in machine learning nor for machine learning experts who aim at testing ideas and models on real-world data in the context of the overall AI system.
We review key efforts made by various AI communities to provide languages for high-level abstractions over learning and reasoning techniques needed for designing complex AI systems. We classify the existing frameworks based on the type of techniques and the data and knowledge representations they use, provide a comparative study of the way they address the challenges of programming real-world applications, and highlight some shortcomings and future directions.
\end{abstract}

\section{Introduction}

\label{Introduction}
While the goal of conventional programming is to automate the tasks that are clearly explainable as a set of step-by-step instructions, the main goal of AI has been to develop programs that make intelligent decisions and solve real-world problems, possibly dealing with ``messy" real world input that could make it difficult to handle using ``conventional" programming. The first AI problem solvers were expert systems that attempted to model the way 
experts reason and make decisions using a set of logical rules. Programming languages like Lisp and Prolog were aimed at making programming for these systems easy even for non-expert users. The idea was to program the domain knowledge in a set of logical rules, and use the rules in a logical reasoning process hidden from the programmers. 

From the traditional AI perspective, this is a declarative programming paradigm in which we program for the {\em what} and not the {\em how}. The expert programs could go beyond an independent set of rules and turn to logical programs with a Turing-complete expressivity, supporting logical inference, for example, by unification and resolution. However, real-world problems are complex and often involve many interdependent parameters. There is a need to interact with naturally occurring data — text, speech, images
and video, streams of financial data, biological sequences — and to reason with respect to complex concepts that often cannot be written explicitly in terms of the raw observed data. It has become clear that formalizing
complex problem solving with a finite set of deterministic logic-based rules is
not possible, nor is it possible to write a conventional structured program, even with a Turing-complete language, for making truly intelligent decisions. Consequently, there has been a rapid paradigm shift from formal modeling to data-driven problem solving. This has affected not only core AI problems like natural language understanding, computer vision, and game playing but also 
real-world problems in many areas including the cognitive sciences, biology, finance, physics, and social sciences. It is becoming more and more common for scientists to think about data-driven solutions using machine learning techniques.

The fundamental goal of machine learning has been to induce programs that can learn from data. However, this 
is not the currently dominating view on machine learning.
In the current view, programs are reduced to models mapping input to output, and learning is an optimization process driven by an objective function of a predefined form.
Thus, machine learning focuses on learning models based on 
classification or regression objective functions rather than generic programs and problem solvers for arbitrary tasks. Nevertheless,
the challenge of "programming" for solving complex tasks and real-world problems using machine learning and intelligent components 
is still very important
but has not been systematically investigated. 
Current machine learning (ML) and AI technologies
do not provide easy ways for domain experts who are not ML/AI experts to develop applications; as we show later, they provide rather cumbersome solutions along multiple dimensions.
Even for AI experts, the new techniques need to be evaluated on messy real-world data rather than well-formed toy problems, which means that anyone who tries to make sense of them -- even an AI expert -- will need to spend a tremendous amount of time and effort due to missing values, formatting errors, anomalies, and other issues. 

Building today's complex AI systems requires extensive programming time and skills in addition to the ability to work with various reasoning and learning paradigms and techniques at a rather low level of abstraction. 
It also requires extensive experimental exploration for model selection, feature selection, and parameter tuning due to lack of theoretical understanding or tools that could be used to abstract over these subtleties. 
Conventional programming languages and software engineering paradigms have 
not been designed to support the challenges faced by users of AI Systems. In particular, they were not designed to deal with messy, real-world data at the appropriate level of abstraction. While this necessitates the use of data-driven methods, programming expert knowledge into current data-driven models  in an intuitive way is highly challenging.
There is a need for innovative paradigms that seamlessly support embedded trainable models, abstracting away most low-level details, and facilitate reasoning with respect to these at the right level of abstraction.  

We believe that this problem is at the root of
many interesting 
and fundamental
research questions and goes beyond 
simply developing good toolboxes and libraries for machine learning based on existing techniques. It requires integrating well-established techniques, dealing with multiple research challenges in data and knowledge representation and reasoning, integration of various machine learning formalisms, innovations in  programming languages and software development.

To help close this gap and facilitate progress in developing what we call here {\em Systems AI},
we survey key 
efforts made in this direction. We emphasize the need to keep some fundamental declarative ideas such as first-order query languages, knowledge representation and reasoning techniques, database management systems (DBMS), and deductive databases (DDB), and attempt to place them within ML formalisms {including classical ML tools, deep learning libraries and automatic differentiation tools}, and integrate them with innovative programming languages and software development techniques, as a way to address complex real-world problems that require both learning and reasoning models. 

%
%
%
%
%

\subsection{AI Application Requirements}

We identify the following criteria as areas of need to enrich existing frameworks with capabilities for {learning-based programming}~\cite{Roth05} and for designing complex AI applications and systems. We also point to a number of questions related to the characteristics of programming languages that enable those requirements. See Figure~\ref{fig:paradigms} for a summary.
\begin{enumerate}

    \item  Easy interaction with raw, heterogeneous data: The key question is \textit{how is communication with the data  performed in the exiting frameworks}?
    
     \item Intuitive means to specify and use domain knowledge:  \textit{What kind of knowledge is needed? Should it be declarative or imperative? How should it be specified?}
      \item  Express uncertainty in data and knowledge: \textit{How should uncertainty be represented? Which underlying formalisms can be used? What kind of expressive power is needed?}
    
    \item Access to various learning, inference and reasoning techniques: \textit{What underlying algorithms are to be supported?}
    
    \item Ability to reuse, combine and chain models and perform flexible inference with complex models/pipelines: \textit{How can we support building end-to-end models from {heterogeneous} components?} 
    
    \item High-level and intuitive abstractions for specifying the requirements of applications: \textit{What should be expressed in a learning-based program? The training objective function? The data? The knowledge? Do we need programs that can learn, or do we need conventional programming that includes learning-based components? Should it be a language or a library? What should be the level of automation? Can we learn the programs automatically?}

\end{enumerate}
  

 
To discuss the existing related paradigms and the key techniques addressing them,
we will use a running example -- designing an intelligent model solving a simple 
entity-mention-relation (EMR) extraction task -- and assume populating a knowledge graph using such information: 

{\bf Given} an input text 
 such as "\emph{Washington works for Associated Press.}", {\bf find} a model that is able to extract the semantic entity types (e.g., people, organizations, and locations) as well as relations between them (e.g., works for, lives in),  and generate the following output: \emph{[Washington]$_{person}$ [works for]$_{worksFor}$ [Associated Press]$_{organization}$}. The chunks of text, along with the labels, will populate a knowledge graph that contains nodes that correspond to entities, and edges that correspond to relations between them. 
 Note that by ``population" we mean that nodes and edges are added for entities and relations, and strings are assigned to these as attributes, identifying the entity type or relation, respectively. 
 
 



%
%
%

\subsection{Related Existing Paradigms} 
\begin{figure}
    \centering
    \includegraphics[width=\textwidth]{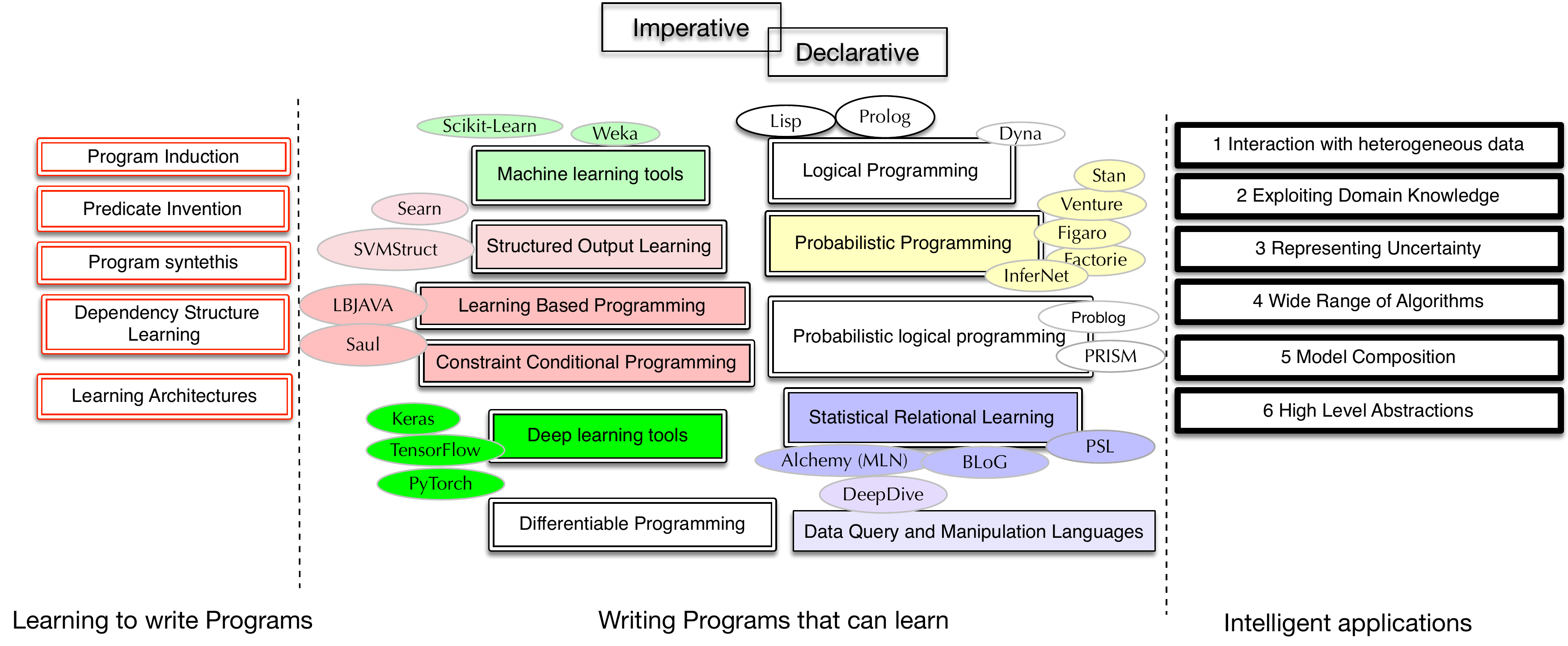}
    \caption{Related paradigms and example frameworks}
    \label{fig:paradigms}
\end{figure}

The AI community has developed various proposals to address the aforementioned requirements for designing intelligent applications.
Indeed, 
there have been various proposals within the AI community 
that address the aforementioned requirements for designing intelligent applications. 
\ignore{Here we point to some of the most well-known and   
most notably Probabilistic (Logic) Programming (P(L)P), Logical Programming (LP), Constrained Conditional Models (CCM) and Statistical Relational Learning/AI (SRL/StarAI). Most of them aim at learning over probabilistic structures and exploiting knowledge in learning. Recently, 
Deep Learning (DL) tools have created easy to use abstractions for programming model configurations of deep neural architectures.}
 We will first review the related communities and some of the frameworks to provide the big picture.  
We will refer back to these frameworks in the following sections when we compare them. The key issues with these frameworks are that: 
\begin{itemize}
\item One still needs deep mastery of ML and AI techniques and methodologies in order to engineer AI systems, and this knowledge far exceeds what most application programmers have. 
\item None of these paradigms covers all the requirements in one unified framework. 
\end{itemize}
Figure~\ref{fig:paradigms} shows a rough picture of various paradigms that are related to learning-based programming in one way or another. The right side shows the six requirements from intelligent applications. In the middle, we point to eight different paradigms, some tightly related, that deal with languages and tools for high-level machine learning and declarative programming. The left side shows concepts related to learning to learn programs. 

\par {Probabilistic Programming Languages (PPLs).} \emph{These languages are designed to describe probabilistic models and perform probabilistic inference}. Given that estimating the parameters of probabilistic models and making predictions based on probabilistic inference is one of the main class of techniques used in machine learning, probabilistic programming languages help users to design and use probabilistic models without worrying about the underlying training and inference algorithms. Examples include Figaro~\cite{pfeffer16}, Venture~\cite{venture}, Stan~\cite{JSSv076i01}, and InferNet~\cite{InferNET}. 

\par{Probabilistic Logic Programming (PLP).} \emph{The aim of these languages is to combine the capacities of probability theory and deductive logic.} When compared to probabilistic programming languages, in addition to the logical reasoning aspect, they bring in capabilities of higher order and compact logical representations of the domain knowledge. The parameters of the PL programs are trained from data, and they can make predictions based on probabilistic logical reasoning. Examples are ProbLog\cite{Raedt07problog:a} and PRISM~\cite{SatoKa97}.

\par{Statistical Relational Learning (SRL).} \emph{This discipline deals with languages that are able to describe complex relational structures and express uncertainty.} 
They do not always rely heavily on logical reasoning but usually exploit a subset of first order logic to express structures. The structures are used during training machine learning models and making inference under uncertainty.
Examples are Markov logic networks~\cite{MLN}, Probabilistic soft logic~\cite{broecheler:uai10} and BLOG (Bayesian Logic)~\cite{MMRSOK05}. 
The relational and logical representations bring in the capabilities of more compact representations, parameter tying and efficient lifted inference~\cite{DeSalvoBraz:2005:LFP:1642293.1642503} in SRL models as well as in probabilistic logical models. 
With a different perspective from these examples, Constrained Conditional Models use the relational representations in learning in the form of logical constraints~\cite{RothYi04,ChangRaRo12}.

\par{Learning-Based Programming.}  \emph{Its special perspective is looking at learning models as first class objects that are able to extract features and make uncertain decisions.} It focuses on the ways that these first class objects can be composed and constrained to form global models to predict complex structures~\cite{Roth05}. The LBJava language~\cite{Rizzolo, RizzoloRo10} and Saul library~\cite{KordjamshidiRoWu15} are based on this idea.    

\par{Classical Machine Learning Toolboxes.} \emph{These are usually libraries designed in general purpose languages and call the training and prediction based on  classical classifiers and regressors}. These cover broad ranges of classification, clustering and regression algorithms that are applied on a form of flat vector representations. Examples are WEKA~\cite{Witten99weka:practical} and Python Scikit-learn libraries\footnote{\url{http://scikit-learn.org/stable/}}, among others.

\par{Structured Learning Tools.} \emph{These tools go beyond classical machine learning toolboxes by allowing the programmer to encode the structure of the multiple output variables and perform inference during training}. SVM-struct\footnote{\url{www.cs.cornell.edu/people/tj/svm_light/svm_struct.html}}, JLIS\footnote{\url {http://cogcomp.org/page/software_view/JLIS}} and SEARN~\cite{DBLP:journals/corr/DaumeLR14} are examples. 

\par{Deep Learning Tools and Languages.} \emph{These are usually libraries within general purpose languages and help with designing deep learning architectures}. Examples are PyTorch\footnote{\url{https://pytorch.org/}} and TensorFlow~\cite{Abadi2016TensorFlowAS} among others.

\par{Differentiable Programming.} \emph{This is a recent paradigm that generalizes deep learning.} 
Imperative programs can be written in terms of differentiable functions where differentials are taken automatically~\cite{Baydin2017AutomaticDI} and the program's parameters are trained/optimized given data to produce appropriate outputs given the inputs. 

\par{Data Query and Manipulation Languages.} 
\emph{Since learning is data-driven, a language for accessing and querying data can be an essential part of a learning-based program.} The ideas in deductive databases~\cite{baranyCKOV17} are relevant as they provide platforms for integration of data and first-order knowledge for inference. The probabilistic databases are also highly related because of their capacity to handle uncertainty in answering database queries and making probabilistic inference{~\cite{Suciu:2011:PD:2031527,denBroeck:2017:QPP:3164891}}. 

We refer back to Figure~\ref{fig:paradigms} when we discuss the existing work in the next section. We connect the notion of writing learning-based programs to that of {\em learning}  learning-based programs, which, in turn, is related to program synthesis, program induction and learning end-to-end differentiable programs. Our goal is to organize the various lines of work related to developing languages for designing machine learning applications and highlight some fundamental research questions that can open new avenues for research on machine learning, programming and developing AI systems.
\ignore{
 We refer back to the parts of Figure~\ref{fig:paradigms} when we discuss the existing work in the next section. We connect the ideas of writing learning-based programs, to learning the learning-based programs which connects to the ideas in program synthesis, program induction and learning end-to-end differentiable programs which is itself a new emerging research --see left side of the Figure. Our goal is to organize the works related to developing languages for designing machine learning applications and highlight some fundamental research questions that can open new avenues for research on machine learning, programming and developing AI systems.}

\section{How do Existing Paradigms Address Application Requirements?} 
\label{requirements}

Given the aforementioned requirements and the key questions to be addressed, we can now explore their relationship with existing frameworks. This allows us 
to discuss the shortcomings of the existing frameworks. We use the EMR example to clarify the concepts when needed. 

\ignore{To discuss aforementioned (1)--(5) as well as
the existing research and key techniques addressing them,
we will use designing an intelligent model solving a simple 
entity-mention-relation (EMR) extraction task as running example: }
\ignore{{\bf Given} an input text 
 as "\emph{Washington works for Associated Press.}", {\bf find} a model that is able to extract the entity types such as people, organizations and locations as well as their relationships, such as, works for, lives in,  and generate the following labels, \emph{[Washington]$_{person}$ [works for]$_{worksFor}$ [Associated Press]$_{organization}$}.}
 

\subsection{Interaction with Heterogeneous Data} 

For real-world applications, organizing and using data is an essential starting point for learning-based programs. For example, in the EMR task, we interact with raw text data (strings). We need to extract useful abstractions from the text
and put raw text into a structure such as a relational database, a knowledge graph or similar data organization for easy access and use in other tasks. We may also want to associate properties of text chunks with them; these could be their semantic types or even a continuous representation (embedding). In this section we point to some of the existing frameworks that facilitate such interactions with both structured and unstructured data in various forms. 

\paragraph{Unstructured data.}
Many real-world systems need to operate on heterogeneous and unstructured data 
such as text and images. To give the data some structure, we need information-extraction modules that could be complex intelligent components themselves. 
{In the EMR task, 
an initial step needed before any semantics is to be inferred via learning components is chunking, that is, splitting the sentence\footnote{Sentence splitting itself is a non-trivial task that may require learning, but in our example, we will assume that sentences are given.} into a number of phrases such as [Washington][Works For][Associated Press][.]. This is a challenging learning task on its own but also provides a primary structure that classifiers can operate on.} Such complex prepossessing steps can also be learned jointly with the main target tasks.

Some of the current research has tried to combine information extraction modules with relational 
DB systems and use standard querying languages for retrieving  information~\cite{Krishnamurthy:2009:SSD:1519103.1519105}. Some 
systems are designed for processing textual data and provide a regular expression interface language to query directly from text~\cite{fextor}. To facilitate working on unstructured data,  efforts have been made to design unified data structures for processing textual data and tools that can operate on those data structures. A well-known example of such a universal data structure is Unstructured Information Management Architecture (UIMA)~\cite{ferrucci2004uima} 
that can be augmented with NLP tools that provide lexical, syntactic and semantic features~\cite{sammons2016edison}.
UIMA focuses on providing a specific internal representation of raw data (in this case, text) but does not support declaring a data model with an arbitrary structure. However, some of the information extraction systems are equipped with very well designed and efficient query languages, such as SystemT~\cite{Krishnamurthy:2009:SSD:1519103.1519105}.

There is a disconnection between such systems and machine learning tools. 
On one hand, such systems do not address learning and inference, \textit{i.e.} their functionality is independent from the design of learning models. However, they could be used as \emph{sensors}\footnote{See~\cite{KordjamshidiRoWu15,RizzoloRo10} for sensor definition.} for information extraction in designing learning models. 
On the other hand, existing machine learning tools do not address the issues involved in  working with unstructured data. Current ML tools such as WEKA~\cite{Witten99weka:practical}  
and newer Python and deep learning packages~\cite{Abadi2016TensorFlowAS} provide easy access to learning algorithms but typically 
 only support flat feature vectors in a specific format,
making it difficult to work with raw or structured data. This is the obvious gap in the existing systems for applying machine learning.

\paragraph{Relational and graph data.}
Many applications require dealing with data with complex structures. Organizing, manipulating and efficient querying from data has been addressed by relational database management systems based on relational representations and standard query languages. These systems are traditionally independent from learning-based models as illustrated by our 
EMR task. We want to learn to extract the entities and relationship and put 
them into a database 
for efficient and easy use. 

Providing ML capabilities on top of relational databases has been followed, for example, in the DeepDive~\cite{zhangRCSWW17}, where first order logical SQL expressions form a Markov logic network or some other relational probabilistic models~\cite{2016Raedt}. These are grounded for parameter estimation and for inference and predictions over relational data. 
In the relational logic-based learning framework of kLog~\cite{frasconi2014klog}, one can use black box classifiers based on relational features which are represented using a logical style and implementations of Datalog. Relational data and relational features can be queried and used directly in machine learning models. The possibility of programming for the objective functions by SQL in DBMS environment and forming learning objectives was followed in the LogicBlox~\cite{arefCGKOPVW15} and RELOOP~\cite{KerstingMT17} systems.
Another example is the Saul~\cite{KordjamshidiRoWu15} language, which is equipped with in-memory graph queries which can be directly used in learning models as features or for constructing learning examples. Moreover, the queries can form global structural patterns between inputs and  outputs~\cite{KKCMSR16,2017_saul_relational_learning_starai}.


One shortcoming which we observed in these frameworks that integrate structured and relational data into learning, is that they cover a specific learning approach and do not provide the flexibility of working with various learning and inference algorithms. They also do not offer flexibility in feature design when working with raw data; in other words, the initial graph/structure should be given to the model. 

\paragraph{Feature extraction.}

One central goal of interaction with data in learning-based programs is to facilitate defining and extracting features from various data sources for learning models. Typically, feature engineering includes a) the ability to obtain low-level sensory 
properties of learning examples (\textit{e.g.}, the length of a phrase or the lemma of its words); b) the capability of selecting, projecting or joining properties for complex and structured data; c) feature induction; d) feature selection; and e) mapping features from one representation to another. This implies that feature extraction is a component that should address the  aforementioned 
issues of interaction with raw data, placing it into structure, and querying the resulting structure. 
Feature extraction approaches can be deterministic, such as 
logical query languages on relational data, or they can be information extraction tools as described above. 
For example, we can place all phrases from the given text in a relational database and then query for all pairs of phrases that have a specific distance between them in the sentence.  

In the NLP domain, Fextor~\cite{fextor} is a tool that provides an internal representation for textual data and provides a library to make queries, like asking the POS-tag 
of a specific word, relying on its fixed internal representation. 
Prior to Fextor, Fex~\cite{CumbyRo03} 
viewed feature extraction from a first order knowledge representation perspective. Their formalization is based on description logic 
where each feature extraction query is answered by logical reasoning. The commonly used machine learning/deep learning libraries provide capabilities for manipulating features as far as those are represented as vectors or matrices (thus no handling of arbitrary structures nor unstructured data) using techniques like dimensionality reduction of convolution operations. 
While there has been research on each of the items (a--e) mentioned above, 
a unifying framework remains elusive, as does a programming environment that facilitates ML. 



\subsection{Exploiting Domain Knowledge}
\label{DomainKnowldge}
We use the term "knowledge" to describe the type of information that goes beyond single data items, is external to the existing data, and expresses relationships between classes of objects in the real world. This is the kind of information that, for example, first order logic formalisms are able to express.   
Different types of domain knowledge can be distinguished based on the type of concepts, functionality or the way those are expressed. In this paper, we classify the type of knowledge based on the latter factor (the way it is expressed from the programming languages' perspective):  \textit{declarative} and \textit{procedural} knowledge.  

{\bf Declarative Knowledge.} Traditional expert systems emphasized the use of  
world knowledge expressed in logical form, due to its declarative nature. 
{Although domain knowledge can convey more information than a set of data items, it is not always straightforward to account for it in classical learning approaches. In our EMR example, while the specific linguistic features of each word/phrase are part of our information about each 
instance, we can have some higher level knowledge over sets of phrases. For example, we know that ``if an arbitrary phrase has type {\em person} it can not be a {\em location}" and that ``if an arbitrary phrase is a person and  another arbitrary phrase is a location the relation between them can not be married". Statistical relational learning models, constrained conditional models \cite{RothYi04,ChangRaRo12}, and probabilistic logical languages~\cite{2016Raedt} address this issue. }
Some 
probabilistic logical models provide the computational models of logical reasoning using resolution, unification or others while the data items are also represented in the same framework, and learning models can be trained based on the data. 
A typical example is Problog~\cite{Raedt07problog:a}. 


{Logical representations of the domain knowledge have been used in other frameworks under the umbrella term of SRL models. These include Constrained Conditional Models~\cite{RothYi04,ChangRaRo12}, Markov Logic Networks~\cite{MLN}, Bayesian Logic~\cite{MMRSOK05}, and Probabilistic Soft Logic models~\cite{broecheler:uai10}. However, the logical representations in these frameworks are usually grounded and generate data instances that form underlying probabilistic graphical models of various kinds. In~\cite{RothYi04}, for example, propositional logical formulas are converted into linear inequalities that constrain the output of structured learning problems.} SRL models do not necessarily consider logical reasoning. Nevertheless, the relational and logical representations provide a compact and expressive mean for higher order information that can potentially be exploited for efficient inference. Representing domain knowledge along with the data has been instantiated in deductive databases such as Datalog~\cite{datalog}, while expressing uncertainties in the data has been considered in probabilistic databases~\cite{2011Suciu}. An example of a deductive database that represents uncertainties, is ProPPR~\cite{Wang:2014:PEF:2908339.2908364},
which has been augmented to learn the probabilities of the facts in the database using neural techniques in TensorLog~\cite{DBLP:journals/corr/CohenYM17}. Learning the structure of SRL models has also been considered\footnote{\url{https://starling.utdallas.edu/software/boostsrl/wiki/}} and shown to be successful in many applications~\cite{natarajanKKS14}. 

{\bf Procedural Knowledge.}  One form of procedural knowledge is knowledge about a specific task that an intelligent agent is supposed to perform. While  knowledge about the data-items, concepts and their relationships is naturally expressed via logical formalisms and in a declarative form, for some domains these representations are less convenient. For example, while the rules of a game (including the legal actions and the goal) could be described in logic, the recipe for cooking a dish or calling a person by phone are inherently procedural and include a sequence of actions.
Depending on the application, we should be able to describe both types of domain knowledge in the learning models. Current programming languages take one of the two mentioned approaches, not both. For example, to program a procedure in Prolog, all of the code still needs to be written in the form of logical rules in a way that the interpreted semantics by Prolog lead to running the intended procedure. This can make writing very simple procedures somewhat unintuitive and difficult to code properly unless the programmer is very experienced with Prolog and its formal semantics.

Using procedural knowledge representations for machine learning can have various interpretations. 
Sometimes, ``imperative programming" refers to the way we express the training and prediction specifications/procedures. However, teaching a machine to perform a task with a sequence of steps may require one to express the procedure of the task as part of the background knowledge. 
The imperative task definition is different from an imperative program that hard codes the objective function of the training. 

For the same reason, in this paper, even defining a task procedure subject to the learning is referred to as ``declaring the procedural domain knowledge". The procedure of a task, expressed in an imperative form, could be taken as the declaration of a specific learning model and be connected to some formal semantics with a different underlying computation from the deterministic sequential execution of a set of instructions. We also call this ``declarative programming" because parts of the domain knowledge are expressed procedurally, but the execution is not deterministic and depends on the trained models. While this might be merely an issue of terminology, we believe this perspective is important to broaden the scope of declarative knowledge representation in the context of learning-based programming. Given this view, we can also call differentiable programs~\cite{bosnjakRNR17} learning-based programs; however, there are severe limitations of what can be expressed in these programs. We will clarify this further when we discuss model composition in Section~\ref{Composition}. An example of an imperative learning based  program  for the EMR task could be a basic if-then-else structure to form a pipeline of decision making. For example, if phrase $x$ is a person then check phrase $y$; if phrase $y$ is a location then check the relationship between $x$ and $y$; and so on. This specifies a procedure for decision making although the decisions are based on learning functions. Nevertheless, it 
guides the formulation of a global objective function for learning. 

\subsection{High-level Abstractions} 
\label{High-level-Abstractions}
 \ignore{this moved to intro: While the goal of conventional programming is to automate tasks that are step-by-step instructions, the main goal of AI 
are programs that make intelligent decisions and solve real-world problems. Recall the very first AI problem solvers. They were expert systems modeling the way experts reason and make decision based on a set of logical rules. Programming languages like Lisp and Prolog made programming them easy even for non-expert users, or at least targeted this idea.

In traditional AI expert systems, the idea has been that only the domain knowledge is programmed in a declarative form in terms of logical facts and rules. The way the rules are used in a logical reasoning and decision making process is hidden from the programmer. This declarative programming paradigm highlights 
the `what' and not the `how'.}


 
Traditional declarative programming often considers programs as the theories of formal logic, but in general, declarative programs could be any high-level specifications of \textit{what} needs to be done where the \textit{how} is left to the language's implementation. All current tools and languages aim at obtaining the right level of abstraction and being declarative in that sense. 
We distinguish between two types of abstractions, a) \emph{data and domain abstractions} and b) \emph{computational and algorithmic abstractions}.

Depending on the type of technique, various abstractions have been made based on both data and computations: 
i) data and domain abstractions in terms of logical model of the domain knowledge, ii) data abstractions based on the dependency structure of variables, iii) computational abstractions based on mathematical functions that form the objective of learning and inference, iv) a combination of data and computational abstractions  representing the model as a procedural partial program. We describe these various perspectives and related implementations. 

Current ML\footnote{\url{scikit-learn.org}} and deep learning tools\footnote{\url{www.tensorflow.org}}$^,$\footnote{\url{openai.com}}$^,$\footnote{\url{https://pytorch.org}} have made a considerable progress towards being more declarative and independent from algorithms, at least for standard ML problems. Using classical ML libraries, the programmer needs to provide feature vectors and to specify only a number of parameters. The programs are written independently from the training algorithms. Retaining the high-level declarations becomes more challenging when the data becomes complex and structured as we go beyond predicting a single variable in the output. We need to use additional domain knowledge beyond data items and feature vectors. 
\paragraph{Logical representation of the domain knowledge.} We have touched on this briefly in Section~\ref{DomainKnowldge} where we described considering domain knowledge in learning.  The paradigms in probabilistic logical programming and statistical relational learning use the idea of representing data and domain abstractions in terms of logical and relational representations. 
\paragraph{Dependency structure of the variables.} Probabilistic programming languages 
facilitate the specifications of (in)dependencies. 
The user declares random variables and their dependency structure 
and other related parameters such as distributions and density functions. 
The structure is specified, used declaratively, and is independent of underlying algorithms for inference and parameter estimation. 
The domain knowledge includes the prior assumptions about the types of the distributions of random variables. 
Reconsider our EMR task. 
We specify the phrases as random variables after we have already obtained an appropriate representation for them. Next, we specify the dependency between each word and its label, or the labels of each word and its adjacent word. Given the data, we can then train the parameters and query 
probabilities of each label or do MAP inference or other computations. Examples of such languages\footnote{\url{probabilistic-programming.org}} are InferNet~\cite{InferNET}, Figaro~\cite{pfeffer16}, AutoBayes~\cite{FischerSc03},  BUGS~\cite{GilksThSp94}, and Stan~\cite{JSSv076i01}. Some of these languages 
are Turing-complete and support
general purpose programs using probabilistic execution traces  (Venture~\cite{venture}, Angelican~\cite{wood-aistats-2014}, Church~\cite{Goodman08church:a}, and Pyro\footnote{\url{pyro.ai}}). The probabilistic logical languages provide another layer of abstraction on top of what probabilistic programming languages already provide. They enable the user to program in terms of data and knowledge and express the dependencies at a logical and conceptual level rather than the (in)dependency structure of the random variables, which is directly used by probabilistic models. The logical representations are given semantics and interpretations that are mapped to lower level probabilistic dependency structures.

\ignore{\paragraph{Logical representation of the domain knowledge.}
Traditional expert systems emphasize  
world knowledge. 
We use the term knowledge for the type of information that goes beyond single data items and expresses the relationships between classes of objects in the real-world. This is the kind of information that, for example, first order logical formalisms are able to express. In our EMR example, while the specific linguistic features of each word/phrase is a part of our information about each 
instance, we can have some higher level knowledge over the sets of phrases. For example, we know that `if an arbitrary phrase is a person it can not be a location' and that `if an arbitrary phrase is a person and another arbitrary phrase is a location the relationship can not be married'. 
Though the domain knowledge can convey more information compared to a set of data items, it is not always straightforward 
to account for it 
in classical learning approaches. Statistical relational learning models and probabilistic logical languages~\cite{2016Raedt} tackle this issue. 
Some 
probabilistic logical models provide the computational models of logical reasoning, using resolution, unification, etc., while the data items are also represented in the same framework, and learning models can be trained based on the data. 
A typical example is Problog~\cite{Raedt07problog:a}. 
Logical representation of the domain knowledge has been used in other frameworks under the umbrella term of SRL models such as MLNs~\cite{MLN} and BLOG~\cite{MMRSOK05}, and PSL models~\cite{broecheler:uai10}. However, the logical representations in these frameworks are all grounded and generate data instances that form underlying probabilistic graphical models of various kinds. In contrast to the above-mentioned probabilistic logical models, the SRL models do not necessarily consider logical reasoning. Representing domain knowledge along with the data has been instantiated in deductive databases such as Datalog ~\cite{datalog} while expressing uncertainties in the data has been considered in probabilistic databases~\cite{2011Suciu}. An example of a deductive database with representing uncertainties is ProPPR~\cite{Wang:2014:PEF:2908339.2908364},
which has been augmented to learn the probabilities of the facts in the database using neural techniques in TensorLog~\cite{DBLP:journals/corr/CohenYM17}.
}

\paragraph{Programming the mathematical objective functions.} Typical examples of this type of abstraction are deep learning tools. The programmer does not specify the structure of the data or the dependencies between variables, but the architecture of the model based on mathematical operators, activation functions, loss functions, etc~\cite{Abadi2016TensorFlowAS}. Given the architecture of the operations, which is a computational graph in contrast to a dependency graph, the program would know how to compute the gradients and what procedure to run for training and prediction. The program specifies the objective function of the training without any concerns about taking the gradients or writing the optimization code. These tools can be seen as specific cases of automated differentiation tools~\cite{Baydin2017AutomaticDI}. If we design the EMR model in this paradigm, we will need to have a vector representation of each phrase beforehand and decide how to represent the structured sentences as vectors or matrices. Deep learning tools will be able to operate on these representations and can help us to specify the architecture of our learning model. We can specify the objective function in terms of multiplications, summations and activation functions. 
The idea of mathematical abstractions has been used in other paradigms, even in probabilistic programming tools such as WOLFE~\cite{riedel2014wolfe}. Such abstractions have been used in the context of designing structured output prediction models such as SSVM\footnote{ \url{www.cs.cornell.edu/people/tj/svm_light/svm_struct.html}} or with search-based inference techniques~\cite{DBLP:journals/corr/DaumeLR14} where the loss and predict procedures can be written in a few lines of code. In SSVM, implementing a task-specific inference algorithm is left to the programmer, while in Searn, a generic search-based algorithm for inference is proposed. The end-to-end program has a sequential and imperative structure rather than a declarative form.

\subsection{Representing Uncertainty} 

Most real data is uncertain due to noise, missing information and/or inherent ambiguities. This has triggered a transition from traditional AI's logical 
 perspective to models that support randomness and probabilities. 
Statistical and probabilistic learning techniques inherently address the issue of uncertainty, and this is reflected in the probabilistic programming and SRL languages~\cite{MLN,MMRSOK05,2016Raedt}. Dealing with uncertainty using probabilistic models has been added to database technology in probabilistic databases~\cite{Suciu:2011:PD:2031527} as well as some deductive databases~\cite{Wang:2014:PEF:2908339.2908364,DBLP:journals/corr/CohenYM17}. It remains a challenging research question to have efficient querying capabilities while dealing with uncertainty in data. 

In real-world scenarios, the uncertainty in data leads to uncertainty in executing tasks. Conventional programming languages by no means address the issue of uncertainty --a main reason why they cannot directly solve real-world problems or facilitate intelligent decision making.  
Uncertainty in a generic problem solving programming paradigm has been addressed in a very limited way. An example of considering uncertainty when programming for problem solving with Turing-complete capabilities can be seen in the implementations of {\it probabilistic logical programming} languages~\cite{Raedt07problog:a,SatoKa97,eisner-2008} as well as {\it probabilistic programming} considering randomness in the execution traces~\cite{Goodman08church:a,venture}. In these frameworks, researchers have used a Turing-complete language in the background, which enables performing any arbitrary task, and have enriched it with uncertainty representation to find the best possible output when lacking evidence for finding the exact output of the program. The uncertainties are interpreted and mapped to a specific formal semantics in the existing languages. In fact, almost all current frameworks use a mapping to a specific type of probabilistic graphical models, therefore, different inference techniques based on various formalisms are often not supported. 


The recent idea of differentiable programming can be seen as a way to deal with uncertainty in procedural programs. The issue of incompleteness is addressed by using a different type of underlying algorithm, typically that of recurrent neural networks and neural Turing machines~\cite{gravesWD14}. 
Based on this type of technique, in~\cite{bosnjakRNR17} for example, the sketch of an imperative program is given while the uncertain components of the program are trained given a set of input/output examples. We believe that there is a need to address the uncertainty and incompleteness in the data and knowledge and, consequently, in executing tasks in their various forms. One should also be able to use various computational models and underlying algorithms which are not limited to solving differentiable objective functions.  


\subsection{Wide Range of Algorithms} 

One of the challenges of designing learning-based AI systems is the lack of theoretical evidence about the effectiveness of various inference and learning approaches. 
This issue limits the programmer to experimentation using trial and error. While automatic exploration is an ideal goal and the first steps have been promising, (see e.g.~\cite{thorntonHHL13,pfefferRK16}), 
connecting representations to a variety of algorithms without much engineering is unexplored. The current programming frameworks mostly support a specific class of algorithms for training and inference. For example, probabilistic programming languages and SRL frameworks are based on inference and learning in probabilistic graphical models (PGM), either 
directed or undirected, or generic factor graphs.
Probabilisitc soft logic considers a PGM too but with more scalable algorithms and more efficient solutions by forming a convex optimization problem in a continuous space for inference. LBJava, RELOOP and Saul map the inference problems under the domain's logical constraints to form integer linear programs and use efficient off-the-shelf techniques in that area to solve inference. In LBJava and Saul, learning independent models offers the opportunity to exploit any arbitrary ML algorithm in the training phase and to perform global 
inference during the prediction phase. The joint training and structured learning is limited and does not cover a variety of techniques. Deep learning tools are also limited to representing differentiable objectives that are optimized based on gradient descent and back-propagation of the errors for training.  

\subsection{Model Composition}
\label{Composition}

As we move towards engineering and using AI systems for increasingly complex real-world tasks, the ability to reuse, combine and chain models, and to perform flexible inference on complex models or pipelines of decision making, becomes an essential issue for learning-based programming. When designing complex models, one key question is how to compose individual models and build more complex ones based on those in the current formalisms. Reconsider our EMR task. We can design a model for classifying entities and another model for classifying the relationships. The final, global EMR model will use them as its building blocks. 

The composition language can be a unified language and consistent with basic ML building block declarations. For example, we can form a global objective using the structured output prediction models and perform collective classification to solve this problem. If we have heterogeneous underlying models based on different techniques, then forming a global objective will not be straightforward, and we will have multiple possibilities for combining models. This issue raises the question of whether the current tools naturally support composition or this should be an additional language on top of forming learning objectives.
If we look at the aforementioned frameworks, we see that the first set of tools for classical ML do not support declarative composition. They rely on the ML and programming expertise of the users to program the model composition imperatively. 

\paragraph{Composition in probabilistic programming.} Probabilistic programming covers the aspect of composition inherently since all involved variables can be declared consistently in one framework and the dependency structure could express the (de)composition semantics for learning and inference. The way that we compose complex models is limited to expressing more global dependencies, and the same dependency structure is used for both training and prediction time. This is not expressive enough for building complex models and pipelines of decision making.
We may need to verify that certain conditions hold and compose arbitrary parts. 

\paragraph{Composition in CCMs.} When designing constrained conditional models (CCMs) in languages such as Saul or LBJava, we need to program the two components of local learning declarations and global constraints specifications. The composition can be done consistently as far as it can be formulated by imposing global constraints and building global models. The current implementations based on CCMs~\cite{KordjamshidiRoWu15,RizzoloRo10} can model pipelines and model composition by considering the learning models as first class objects where their outputs can be used to form new learners and new layers of abstractions.   In the frameworks that are designed as libraries of general-purpose languages, the compositions can be made by the programmer although we believe a composition language with well-defined semantics will provide a better way to design complex models with explicit structure, end-to-end.  

\paragraph{Composition in deep neural models.} The deep learning tools rely on general purpose programming environments and the ML and programming skills of users to compose models imperatively. They provide a way to design single models, though CapsNets~\cite{sabourFH17} recently made a first step towards learning compositions in deep networks. {Designing Neural Module Networks is another related direction to build neural modules per domain concepts and compose them explicitly and dynamically for language and vision understanding~\cite{Andreas2016NeuralMN}. These models, though modular and composable,  still rely on end-to-end neural training based on continuous representations.}

\paragraph{Programs seen as compositions of models.} While the composition of the trained models is helpful in designing and programming complex models, one new issue arises. Can we parameterize the programs that include learning-based components and in turn learn the composition itself?
This is a less established line of research. It is not clear how the structure of the program can be represented or what the parameters of the program will be. 
Differentiable programs could be seen as an important step in this direction. They are 
related to program synthesis in the sense that we learn a program from inputs and outputs. They are also related to the idea of learning the parameters of the composition of learning-based components. We can provide the structure or the scheme of the program and learn parts of it. 
The latter perspective should be distinguished from program synthesis 
because the target programs that we learn are not deterministic programs like sorting. They are never complete and only estimate an output given a partial structure and make inferences. They are unlikely to be learnable in a full deterministic structure. 


\section{Declarative Learning-based Programming: An Integration Paradigm}
\begin{figure}
    \centering
    \includegraphics[width=\textwidth]{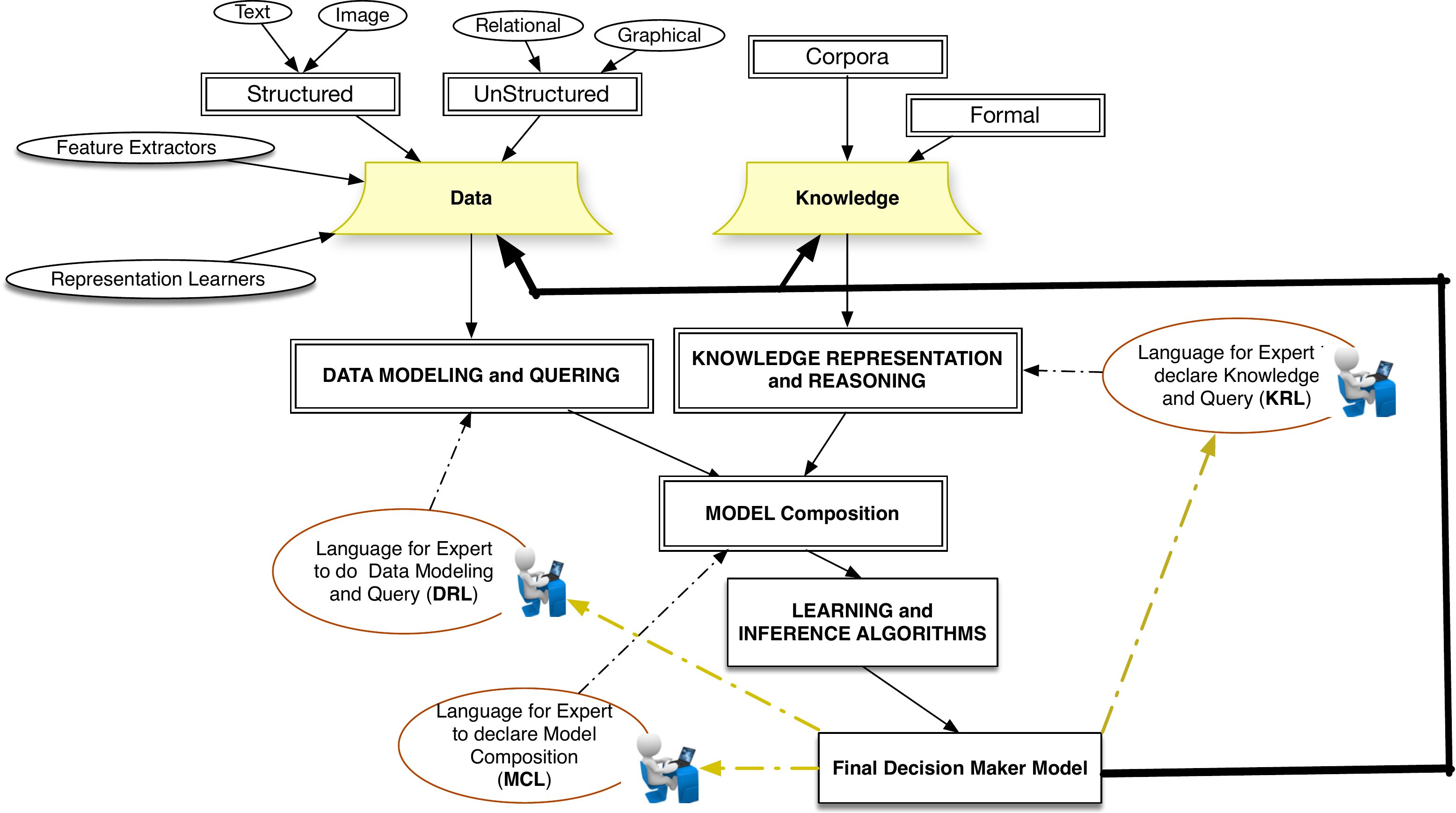}
    \caption{The main components of a learning-based programming language.}
    \label{fig:overview}
\end{figure}

While existing paradigms does address some of the capabilities (1)-(6) from the introduction, there is still a need to integrate more aspects to design truly smart AI systems that 
can constantly collect weak signals
independently of specific tasks and relate them on the fly to solve a task without supervision~\cite{roth17}.

We argue here for a paradigm that highlights the aspect of \emph{learning from data} as a pivot, tries to extend the capabilities of designing intelligent systems around this concept, and addresses the above-mentioned challenges accordingly by allowing programming to construct complex configurations using basic learning building blocks. 
Figure~\ref{fig:overview} shows a rough sketch of a platform that can address the application requirements from an AI-systems development perspective. The platform integrates the capabilities for working with heterogeneous data and knowledge from various resources. This implies that there will be a need for data modeling and representation language (DLR), knowledge representation language (KRL), and model composition language (MCL). These three languages should have access to a set of learning and inference algorithms and allow for domain experts to design models interactively. The output of the intelligent models is either new data or new knowledge that is added back into an evolving intelligent system. 
In the previous sections, we argue for the need for new abstractions with the following characteristics to facilitate programming to interact with AI-systems. 

\paragraph{Abstractions that are independent from computations.}
The idea of learning-based programming, which has been initiated in~\cite{Roth05}, supports the necessity of data abstractions, hiding the algorithmic details and even hiding high-level algorithmic abstractions. Learning is mapping from one data abstraction layer to another given the data instances, starting from raw data. The user needs to specify the intended abstractions for an application in hand, and the system should figure out how to perform the actual mappings. While this abstraction is similar to the argument for in logical formalisms, here we are not limited to logical predicates. Our primitives can be concept-learners that are represented by arbitrary functions. The mapping computations are not limited to logical reasoning mechanisms, and heterogeneous learners can take the data and learn the mappings. LBJava~\cite{Rizzolo11} was a first implementation of this idea, based on the CCM~\cite{ChangRaRo12} computational model. Learners are first class objects, and the domain knowledge also represented in terms of data abstractions can be used by learners to make global and consistent mappings.
RELOOP~\cite{KerstingMT17} took a similar approach from a mathematical programming perspective, combining relational and mathematical programming aspects embedded within an imperative language (in this case, Python). Saul~\cite{KordjamshidiRoWu15} has been proposed with a similar computational model and the possibility of joint training of global models. Saul is in the form of a library without the data-driven compilation step, and it comes with explicit support for the representation of the data as a graph for relational feature engineering. 
The data graph representation helps to specify domain concepts and their relationships. Some concepts are connected to sensors abstracting away from raw data and some are concept learners. 

\paragraph{Abstractions that facilitate algorithmic coverage.}
Most of the frameworks mentioned in the previous sections have  limited coverage of algorithms. While some of these are more flexible than others in supporting heterogeneous computational building blocks, training complex configurations with global learning is addressed with one class of techniques in each framework such as probabilistic inference, integer linear programming, or dynamic programming and search. When the learning model abstractions are based on the data abstractions in addition to domain knowledge and generic problem specification, in principle, these representations can be connected to various computational models. In contrast, the representations based on computational abstractions (such as deep learning methods) are more bounded to the type of underlying techniques for computations {and impose more limitations on algorithmic coverage.}  

\paragraph{Abstractions that help in closing the loop of data to knowledge.} Intelligent systems need to evolve over time. As they receive more data and knowledge, they find better abstractions of the data, as illustrated by NELL~\cite{mitchellCHTBCMG15}, Never-Ending Language Learning. Representations of the learning models based on the data and knowledge  
will naturally support feeding the current models (which will be trained concepts) to obtain new abstraction layers. This will also naturally support the model composition. There will be a direct connection between how we compose models and how we compose real-world concepts. Such abstractions will help to close the loop of moving from data to knowledge and exploiting it to generate new concepts. 

\paragraph{Abstractions that help with learning the programs.}  
While the goal of ML is to write programs that can learn to do a task or make a decision, a more ambitious goal would be to learn the structure of the programs from the data. From the classical ML perspective, this is connected to structure learning. An example is learning the dependency structure of the probabilistic models such as Bayesian networks, see~\cite{kollerFreidman2009}. Another dimension of the problem is learning features or feature induction, which has been investigated in the classical machine learning community for years~\cite{Molina:2002:FSA:844380.844722}. Learning global constraints by analyzing the data is another line of work leading towards learning the structure of the learning models~\cite{bessiereRGKNNOP17}, which is related to traditional rule learning models that can guide the issues related to learning the programs.  

In the programming languages community, this problem is very close to program induction from the inputs and outputs. 
From the classical AI perspective, this is also related to inductive logic programming and program induction (see e.g.~\cite{muggletonR94}). 
These ideas go beyond learning propositionalized rules 
and are about learning logical programs. They can be seen as a set of rules augmented by global formal semantics for unification, entailment, and will be treated together rather than independently.
In fact, inductive logic programming can be considered to be at the same level as other learning algorithms where the structure of the model (i.e., the program) is already given. This structure is usually in the form of a language bias, which is very different from the way the model's structure is defined in (non-relational) statistical learning paradigms. The ultimate case of program induction, learning the programs in the framework of logic, is when the domain predicates are not fully specified but need to be invented during learning~\cite{stahl95}. 
A logical programming language or a classical general purpose programming language, even if it is Turing equivalent, will not be able to solve an AI-complete problem. 
If we represent the structure of a learning model with these paradigms, we still need to think about the parameters addressing 
incompleteness and uncertainty for solving problems intelligently. {This is the major difference between the work done in the scope of program synthesis compared to learning programs that are intended for AI research.}

{Our choice of program content and representation, discussed in Section~\ref{High-level-Abstractions}, is a key factor that influences the way we think about learning the programs themselves and the types of techniques that will be developed in this direction. Depending on the representation of the programs, learning programs can involve learning deep architectures, learning dependence structures or learning classic machine learning features.} 



\paragraph{Other issues from AI and learning-based systems perspective.}
The present paper focuses on the issues related to appropriate and easy-to-use abstractions and coverage of various formalisms for learning-based programs. It does not tackle the many issues to be addressed for the AI systems~\cite{Stoica:EECS-2017-159} that will serve as platforms for these declarative languages. We need to solve similar problems as database management systems when designing AI systems. It is imperative to have learning-based management systems that can deal with security and privacy of data as well as learning models, scalability of learning and inference, distributed and parallel implementations, concurrency and more. There are new issues such as fairness and explainability to be addressed in AI and learning-based management systems.

\section{Conclusion}


Triggered by the emerging research area of Systems AI---the computational and mathematical modeling of complex AI systems---we provided an overview 
on 
{\it declarative learning-based programming} languages as a central component of such a mission and as an interface to interact with AI systems for designing, training and using them for real-world decision-making and task performance. We discussed the related works that can help to design such a language covering a) the type of abstraction that they make over the data and computations, b) the type of techniques that they cover for learning and reasoning/inference c) the way they address the interaction with data and the issue of incompleteness and uncertainty d) the way that those facilitate designing complex models by composition of simpler models. More importantly, we reviewed 
Finally, we emphasized that working on the declarative programming languages that describe the programs in terms of data, knowledge and declaring task procedures will pave the way for training AI systems by natural interactions ~\cite{gluck2018interactive}. The declarative programs can be seen as intermediate representations that intelligent systems can receive directly from the programmers, or ideally learn/infer them from natural interactions in the real world. 

\bibliographystyle{plain}
\bibliography{cited,ccg,new,allpapers,LBP_references,ijcai18}

\begin{thebibliography}{10}

\bibitem{Abadi2016TensorFlowAS}
Martin Abadi, Paul Barham, Jianmin Chen, Zhifeng Chen, Andy Davis, Jeffrey
  Dean, Matthieu Devin, Sanjay Ghemawat, Geoffrey Irving, Michael Isard,
  Manjunath Kudlur, Josh Levenberg, Rajat Monga, Sherry Moore, Derek~Gordon
  Murray, Benoit Steiner, Paul~A. Tucker, Vijay Vasudevan, Pete Warden, Martin
  Wicke, Yuan Yu, and Xiaoqiang Zhang.
\newblock Tensorflow: A system for large-scale machine learning.
\newblock In {\em OSDI}, 2016.

\bibitem{Andreas2016NeuralMN}
Jacob Andreas, Marcus Rohrbach, Trevor Darrell, and Dan Klein.
\newblock Neural module networks.
\newblock {\em 2016 IEEE Conference on Computer Vision and Pattern Recognition
  (CVPR)}, pages 39--48, 2016.

\bibitem{arefCGKOPVW15}
Molham Aref, Balder ten Cate, Todd~J. Green, Benny Kimelfeld, Dan Olteanu, Emir
  Pasalic, Todd~L. Veldhuizen, and Geoffrey Washburn.
\newblock Design and implementation of the {LogicBlox} system.
\newblock In {\em Proc. of the 2015 {ACM} {SIGMOD} International Conference on
  Management of Data}, pages 1371--1382, 2015.

\bibitem{baranyCKOV17}
Vince B{\'{a}}r{\'{a}}ny, Balder ten Cate, Benny Kimelfeld, Dan Olteanu, and
  Zografoula Vagena.
\newblock Declarative probabilistic programming with datalog.
\newblock {\em {ACM} Trans. Database Syst.}, 42:22:1--22:35, 2017.

\bibitem{Baydin2017AutomaticDI}
Atilim~Gunes Baydin, Barak~A. Pearlmutter, and Alexey Radul.
\newblock Automatic differentiation in machine learning: a survey.
\newblock {\em Journal of Machine Learning Research}, 18:153:1--153:43, 2017.

\bibitem{bessiereRGKNNOP17}
Christian Bessiere, Luc~De Raedt, Tias Guns, Lars Kotthoff, Mirco Nanni,
  Siegfried Nijssen, Barry O'Sullivan, Anastasia Paparrizou, Dino Pedreschi,
  and Helmut Simonis.
\newblock The inductive constraint programming loop.
\newblock {\em {IEEE} Intelligent Systems}, 32(5):44--52, 2017.

\bibitem{bosnjakRNR17}
Matko Bo{\v{s}}njak, Tim Rockt{\"a}schel, Jason Naradowsky, and Sebastian
  Riedel.
\newblock Programming with a differentiable {F}orth interpreter.
\newblock In {\em Proc. of 34th ICML}, pages 547--556, 2017.

\bibitem{fextor}
Bartosz Broda, Pawe{\l} Kedzia, Micha{\l} Marci{\'n}czuk, Adam Radziszewski,
  Rados{\l}aw Ramocki, and Adam Wardy{\'n}ski.
\newblock Fextor: A feature extraction framework for natural language
  processing: A case study in word sense disambiguation, relation recognition
  and anaphora resolution.
\newblock In {\em Computational Linguistics}, pages 41--62. Springer, 2013.

\bibitem{broecheler:uai10}
Matthias Broecheler, Lilyana Mihalkova, and Lise Getoor.
\newblock Probabilistic similarity logic.
\newblock In {\em Conference on Uncertainty in Artificial Intelligence}, 2010.

\bibitem{JSSv076i01}
Bob Carpenter, Andrew Gelman, Matthew Hoffman, Daniel Lee, Ben Goodrich,
  Michael Betancourt, Marcus Brubaker, Jiqiang Guo, Peter Li, and Allen
  Riddell.
\newblock Stan: A probabilistic programming language.
\newblock {\em Journal of Statistical Software, Articles}, 76(1):1--32, 2017.

\bibitem{ChangRaRo12}
Ming-Wei Chang, Lev Ratinov, and Dan Roth.
\newblock Structured learning with constrained conditional models.
\newblock {\em Machine Learning}, 88(3):399--431, 6 2012.

\bibitem{DBLP:journals/corr/CohenYM17}
William~W. Cohen, Fan Yang, and Kathryn Mazaitis.
\newblock Tensorlog: Deep learning meets probabilistic dbs.
\newblock {\em CoRR}, abs/1707.05390, 2017.

\bibitem{CumbyRo03}
Chad Cumby and Dan Roth.
\newblock On kernel methods for relational learning.
\newblock In {\em Proceedings of the International Conference on Machine
  Learning (ICML)}, 2003.

\bibitem{DBLP:journals/corr/DaumeLR14}
Hal {Daum{\'{e}} III}, John Langford, and St{\'{e}}phane Ross.
\newblock Efficient programmable learning to search.
\newblock {\em CoRR}, abs/1406.1837, 2014.

\bibitem{2016Raedt}
Luc {De Raedt}, Kristian Kersting, Sriraam Natarajan, and David Poole.
\newblock {\em Statistical Relational Artificial Intelligence: Logic,
  Probability, and Computation}.
\newblock Morgan {\&} Claypool Publishers, 2016.

\bibitem{Raedt07problog:a}
Luc {De Raedt}, Angelika Kimmig, and Hannu Toivonen.
\newblock Problog: a probabilistic {P}rolog and its application in link
  discovery.
\newblock In {\em Proceedings of the 20th International Joint Conference on
  Artificial Intelligence}, pages 2468--2473. AAAI Press, 2007.

\bibitem{DeSalvoBraz:2005:LFP:1642293.1642503}
Rodrigo De~Salvo~Braz, Eyal Amir, and Dan Roth.
\newblock Lifted first-order probabilistic inference.
\newblock In {\em Proceedings of the 19th International Joint Conference on
  Artificial Intelligence}, IJCAI'05, pages 1319--1325, San Francisco, CA, USA,
  2005. Morgan Kaufmann Publishers Inc.

\bibitem{denBroeck:2017:QPP:3164891}
Guy~Van den Broeck and Dan Suciu.
\newblock {\em Query Processing on Probabilistic Data: A Survey}.
\newblock Now Publishers Inc., Hanover, MA, USA, 2017.

\bibitem{MLN}
Perdo Domingos and Matthew Richardson.
\newblock Markov logic: A unifying framework for statistical relational
  learning.
\newblock In {\em {ICML}'04 Workshop on Statistical Relational Learning and its
  Connections to Other Fields}, pages 49--54, 2004.

\bibitem{eisner-2008}
Jason Eisner.
\newblock Dyna: {A} {\em non}-probabilistic programming language for
  probabilistic {AI}.
\newblock Extended abstract for talk at the NIPS*2008 Workshop on Probabilistic
  Programming, 2008.

\bibitem{ferrucci2004uima}
David Ferrucci and Adam Lally.
\newblock {UIMA}: an architectural approach to unstructured information
  processing in the corporate research environment.
\newblock {\em Natural Language Engineering}, 10(3-4):327--348, 2004.

\bibitem{FischerSc03}
Bernd Fischer and Johann Schumann.
\newblock Autobayes: a system for generating data analysis programs from
  statistical models.
\newblock {\em Journal of Functional Programming}, 13(3):483--508, 2003.

\bibitem{frasconi2014klog}
Paolo Frasconi, Fabrizio Costa, Luc De~Raedt, and Kurt De~Grave.
\newblock klog: A language for logical and relational learning with kernels.
\newblock {\em Artificial Intelligence}, 217:117--143, 2014.

\bibitem{GilksThSp94}
Walter~R. Gilks, Avis Thomas, and David~J. Spiegelhalter.
\newblock A language and program for complex bayesian modeling.
\newblock {\em The Statistician}, 43(1):169--177, 1994.

\bibitem{gluck2018interactive}
Kevin~A. Gluck, John Laird, and Julia Lupp.
\newblock {\em Interactive Task Learning: Humans, Robots, and Agents Acquiring
  New Tasks Through Natural Interactions}.
\newblock Str{\"u}ngmann Forum reports. MIT Press, 2018.

\bibitem{Goodman08church:a}
Noah~D. Goodman, Vikash~K. Mansinghka, Daniel~M. Roy, Keith Bonawitz, and
  Joshua~B. Tenenbaum.
\newblock Church: A language for generative models.
\newblock In {\em In UAI}, pages 220--229, 2008.

\bibitem{datalog}
Georg Gottlob, Stefan Ceri, and Letizia Tanca.
\newblock What you always wanted to know about {Datalog} (and never dared to
  ask).
\newblock {\em IEEE Transactions on Knowledge \& Data Engineering}, 1:146--166,
  03 1989.

\bibitem{gravesWD14}
Alex Graves, Greg Wayne, and Ivo Danihelka.
\newblock Neural turing machines.
\newblock {\em CoRR}, abs/1410.5401, 2014.

\bibitem{KerstingMT17}
Kristian Kersting, Martin Mladenov, and Pavel Tokmakov.
\newblock Relational linear programming.
\newblock {\em Artif. Intell.}, 244, 2017.

\bibitem{kollerFreidman2009}
Daphne Koller and Nir Friedman.
\newblock {\em Probabilistic Graphical Models - Principles and Techniques}.
\newblock {MIT} Press, 2009.

\bibitem{KKCMSR16}
Parisa Kordjamshidi, Daniel Khashabi, Christos Christodoulopoulos, Bhargav
  Mangipudi, Sameer Singh, and Dan Roth.
\newblock Better call saul: Flexible programming for learning and inference in
  nlp.
\newblock In {\em Proc. of the International Conference on Computational
  Linguistics (COLING)}, 2016.

\bibitem{2017_saul_relational_learning_starai}
Parisa Kordjamshidi, Sameer Singh, Daniel Khashabi, Christos
  Christodoulopoulos, Mark Summons, Saurabh Sinha, and Dan Roth.
\newblock Relational learning and feature extraction by querying over
  heterogeneous information networks.
\newblock In {\em Seventh International Workshop on Statistical Relational AI
  (StarAI)}, 2017.

\bibitem{KordjamshidiRoWu15}
Parisa Kordjamshidi, Hao Wu, and Dan Roth.
\newblock Saul: Towards declarative learning based programming.
\newblock In {\em Proc. of the International Joint Conference on Artificial
  Intelligence (IJCAI)}, 7 2015.

\bibitem{Krishnamurthy:2009:SSD:1519103.1519105}
Rajasekar Krishnamurthy, Yunyao Li, Sriram Raghavan, Frederick Reiss,
  Shivakumar Vaithyanathan, and Huaiyu Zhu.
\newblock Systemt: A system for declarative information extraction.
\newblock {\em SIGMOD Rec.}, 37(4):7--13, March 2009.

\bibitem{venture}
Vikash~K. Mansinghka, Daniel Selsam, and Yura~N. Perov.
\newblock Venture: a higher-order probabilistic programming platform with
  programmable inference.
\newblock {\em CoRR}, abs/1404.0099, 2014.

\bibitem{MMRSOK05}
Brian Milch, Bhaskara Marthi, Stuart Russell, David Sontag, Daniel~L. Ong, and
  Andrey Kolobov.
\newblock {BLOG}: Probabilistic models with unknown objects.
\newblock In {\em Proceedings of the International Joint Conference on
  Artificial Intelligence (IJCAI)}, 2005.

\bibitem{InferNET}
Tom Minka, John~M. Winn, John~P. Guiver, and David~A. Knowles.
\newblock {Infer.NET 2.5}, 2012.
\newblock Microsoft Research Cambridge. http://research.microsoft.com/infernet.

\bibitem{mitchellCHTBCMG15}
Tom~M. {Mitchell {\it et al.}}
\newblock Never-ending learning.
\newblock In {\em Proc. of the Twenty-Ninth {AAAI} Conference on Artificial
  Intelligence (AAAI)}, pages 2302--2310, 2015.

\bibitem{Molina:2002:FSA:844380.844722}
Luis~Carlos Molina, Llu\'{\i}s Belanche, and \`{A}ngela Nebot.
\newblock Feature selection algorithms: A survey and experimental evaluation.
\newblock In {\em Proceedings of the 2002 IEEE International Conference on Data
  Mining}, ICDM '02, pages 306--, Washington, DC, USA, 2002. IEEE Computer
  Society.

\bibitem{muggletonR94}
Stephen Muggleton and Luc {De Raedt}.
\newblock Inductive logic programming: Theory and methods.
\newblock {\em J. Log. Program.}, 19/20:629--679, 1994.

\bibitem{natarajanKKS14}
Sriraam Natarajan, Kristian Kersting, Tushar Khot, and Jude~W. Shavlik.
\newblock {\em Boosted Statistical Relational Learners - From Benchmarks to
  Data-Driven Medicine}.
\newblock Springer Briefs in Computer Science. Springer, 2014.

\bibitem{pfeffer16}
Avi Pfeffer.
\newblock {\em Practical Probabilistic Programming}.
\newblock Manning Publications, 2016.

\bibitem{pfefferRK16}
Avi Pfeffer, Brian~E. Ruttenberg, and William Kretschmer.
\newblock Structured factored inference: {A} framework for automated reasoning
  in probabilistic programming languages.
\newblock {\em CoRR}, abs/1606.03298, 2016.

\bibitem{riedel2014wolfe}
S.~Riedel, S.~Singh, V.~Srikumar, T.~Rockt{\"a}schel, L.~Visengeriyeva, and
  J.~Noessner.
\newblock {WOLFE}: strength reduction and approximate programming for
  probabilistic programming.
\newblock {\em Statistical Relational Artificial Intelligence}, 2014.

\bibitem{Rizzolo}
N.D. Rizzolo.
\newblock {\em Learning based programming}.
\newblock PhD thesis, University of Illinois at Urbana-Champaign, 2011.

\bibitem{Rizzolo11}
Nicholas Rizzolo.
\newblock {\em Learning Based Programming}.
\newblock PhD thesis, 2011.

\bibitem{RizzoloRo10}
Nicholas Rizzolo and Dan Roth.
\newblock Learning based java for rapid development of {NLP} systems.
\newblock In {\em Proc. of the International Conference on Language Resources
  and Evaluation (LREC)}, Valletta, Malta, 5 2010.

\bibitem{RothYi04}
D.~Roth and W.~Yih.
\newblock A linear programming formulation for global inference in natural
  language tasks.
\newblock In Hwee~Tou Ng and Ellen Riloff, editors, {\em Proc. of the
  Conference on Computational Natural Language Learning (CoNLL)}, pages 1--8.
  Association for Computational Linguistics, 2004.

\bibitem{Roth05}
Dan Roth.
\newblock Learning based programming.
\newblock {\em Innovations in Machine Learning: Theory and Applications}, 2005.

\bibitem{roth17}
Dan Roth.
\newblock Incidental supervision: Moving beyond supervised learning.
\newblock In {\em Proc. of the Conference on Artificial Intelligence (AAAI)}, 2
  2017.

\bibitem{sabourFH17}
Sara Sabour, Nicholas Frosst, and Geoffrey~E. Hinton.
\newblock Dynamic routing between capsules.
\newblock In {\em Proc. of the Annual Conference on Neural Information
  Processing Systems (NIPS)}, pages 3859--3869, 2017.

\bibitem{sammons2016edison}
Mark Sammons, Christos Christodoulopoulos, Parisa Kordjamshidi, Daniel
  Khashabi, Vivek Srikumar, and Dan Roth.
\newblock Edison: Feature extraction for nlp, simplified.
\newblock In {\em LREC}, 2016.

\bibitem{SatoKa97}
Taisuke Sato and Yoshitaka Kameya.
\newblock Prism: A language for symbolic-statistical modeling.
\newblock In {\em Proceedings of the International Joint Conference on
  Artificial Intelligence (IJCAI)}, pages 1330--1339, 1997.

\bibitem{stahl95}
Irene Stahl.
\newblock The appropriateness of predicate invention as bias shift operation in
  {ILP}.
\newblock {\em Machine Learning}, 20(1/2):95--117, 1995.

\bibitem{Stoica:EECS-2017-159}
Ion Stoica, Dawn Song, Raluca~Ada Popa, David~A. Patterson, Michael~W. Mahoney,
  Randy~H. Katz, Anthony~D. Joseph, Michael Jordan, Joseph~M. Hellerstein,
  Joseph Gonzalez, Ken Goldberg, Ali Ghodsi, David~E. Culler, and Pieter
  Abbeel.
\newblock A {B}erkeley view of systems challenges for {AI}.
\newblock Technical Report UCB/EECS-2017-159, EECS Department, University of
  California, Berkeley, Oct 2017.

\bibitem{Suciu:2011:PD:2031527}
Dan Suciu, Dan Olteanu, R.~Christopher, and Christoph Koch.
\newblock {\em Probabilistic Databases}.
\newblock Morgan \& Claypool Publishers, 1st edition, 2011.

\bibitem{2011Suciu}
Dan Suciu, Dan Olteanu, Christopher R{\'{e}}, and Christoph Koch.
\newblock {\em Probabilistic Databases}.
\newblock Morgan {\&} Claypool Publishers, 2011.

\bibitem{thorntonHHL13}
Chris Thornton, Frank Hutter, Holger~H. Hoos, and Kevin Leyton{-}Brown.
\newblock Auto-weka: combined selection and hyperparameter optimization of
  classification algorithms.
\newblock In {\em Proc. of the 19th {ACM} {SIGKDD} International Conference on
  Knowledge Discovery and Data Mining (KDD)}, pages 847--855, 2013.

\bibitem{Wang:2014:PEF:2908339.2908364}
William~Yang Wang, Kathryn Mazaitis, and William~W. Cohen.
\newblock Proppr: Efficient first-order probabilistic logic programming for
  structure discovery, parameter learning, and scalable inference.
\newblock In {\em Proc. of the 13th AAAI Conference on Statistical Relational
  AI}, AAAIWS'14-13, pages 133--134. AAAI Press, 2014.

\bibitem{Witten99weka:practical}
Ian~H. Witten, Eibe Frank, Len Trigg, Mark Hall, Geoffrey Holmes, and Sally~Jo
  Cunningham.
\newblock Weka: Practical machine learning tools and techniques with java
  implementations, 1999.

\bibitem{wood-aistats-2014}
Frank Wood, Jan~Willem van~de Meent, and Vikash Mansinghka.
\newblock A new approach to probabilistic programming inference.
\newblock In {\em Proc. of the 17th International conference on Artificial
  Intelligence and Statistics}, pages 1024--1032, 2014.

\bibitem{zhangRCSWW17}
Ce~Zhang, Christopher R{\'{e}}, Michael~J. Cafarella, Jaeho Shin, Feiran Wang,
  and Sen Wu.
\newblock Deepdive: declarative knowledge base construction.
\newblock {\em Commun.~{ACM}}, 60(5):93--102, 2017.

\end{thebibliography}
\end{document}